\documentclass[preprint,12pt]{elsarticle}

\usepackage{amssymb}
\usepackage{amsmath,graphicx,cite}
\usepackage{epsfig,amssymb}
\usepackage{epstopdf}
\usepackage{xcolor}
\usepackage{algorithmic}
\usepackage{algorithm}

\journal{ }

\begin{document}

\begin{frontmatter}



\title{An Adaptive Task-Related Component Analysis Method for SSVEP recognition}


\author[inst1]{Vangelis P. Oikonomou}

\affiliation[inst1]{organization={Information Technologies Institute, Centre for Research and Technology Hellas},
            addressline={Thermi-Thessaloniki}, 
            city={Thessaloniki},
            postcode={57001}, 
            country={Greece}}

\begin{abstract}
Steady-state visual evoked potential (SSVEP) recognition methods are equipped 
with learning from the subject’s calibration data, and they can achieve extra 
high performance in the SSVEP-based brain–computer interfaces (BCIs), however their 
performance deteriorate drastically if the calibration trials are insufficient. 
This study develops a new method to learn from limited calibration data
and it proposes and evaluates a novel adaptive data-driven spatial 
filtering approach for enhancing SSVEPs
detection. The spatial filter learned from each stimulus utilizes temporal 
information from the corresponding EEG trials. To introduce the temporal 
information into the overall procedure, an multitask learning approach, 
based on the bayesian framework, is adopted. The performance of the proposed 
method was evaluated into two publicly available benchmark datasets, and 
the results demonstrated that our method outperform competing methods by a 
significant margin. 
\end{abstract}




\end{frontmatter}


\section{Introduction}

A Brain Computer Interface (BCI) system is a device that it translates 
human brain activity into artificially generated control signals,
providing us with an alternative communication medium, other 
than physical communication, which can be used to help people with motor
disabilities \citep{Wolpaw:2002}, to augment communication abilities of healthy
individuals,\citep{Alonso:2012}, for entertainment \citep{Alonso:2012}, and,
for neuromarketing purposes\citep{Kalaganis:2021}.
Brain activity can be measured with various specialized devices such as MRI scanners
and electroencephalograms (EEG)-based devices, where, EEG devices is widely used
since the required equipment is simple and inexpensive. 
EEG-based BCI systems utilize various brain responses such as motor imagery and visual 
responses, from which, the use of Steady State Visual Evoked Potentials (SSVEPs) 
have attracted the attention of many researchers due to their lower training 
requirements for the end-user and higher information 
transfer rates \citep{Bin:2009,Zerafa:2018}.  
When an individual is looking into a visual stimulus, which is flashing as a fixed frequency,
then a brain response is revealed in occipital and occipital - parietal areas of individual's
brain which is called SSVEP response\citep{Gao:2014}. 
A SSVEP response contains sinusoidal components which are related to the fundamental 
frequency of the visual stimulus as well as its harmonics. The overarching goal of a
SSVEP BCI system is to detect the different 
frequency components corresponding to the visual stimuli 
and translate them into commands, by using an 
EEG-based pattern recognition algorithms. 

The recognition of SSVEP responses involves the use of Machine Learning 
(ML) algorithms. 
Linear classifiers such as Support Vector
Machines (SVMs) and the Linear Discriminant Analysis (LDA) 
have been used to detect SSVEPs \citep{Oikonomou:2016}. 
In addition, in \citep{Wang:2016} the use of Multivariate Linear 
Regression (MLR) was  proposed to learn discriminative features for 
improving SSVEP classification, while, in \citep{Oikonomou:2019jbhi} 
kernel - based extensions of MLR were proposed using SSVEP-related kernels
as an integral part of the Sparse Bayesian Learning (SBL) framework.
Furthermore, Deep Learning (DL) approaches using 
Convolutional Neural Networks (CNN), based on time frequency analysis, 
are used to discriminate SSVEP responses \citep{Cecotti:2011, Kwak:2017}. 

However, SSVEP responses present specific frequency and spatial 
characteristics, hence methods utilizing these characteristics have been proposed. More specifically, methods based on Power Spectrum 
Density Analysis (PSDA) were widely used for frequency detection
\citep{Wang:2010}, where the target frequency is assigned to
the frequency corresponding to maximum value of PSD. 
However, the efficient calculation of PSD requires a relatively large
time window, and it is sensitive to noise \citep{Lin:2006,Friman:2007}.
To avoid the above shortcomings of PSDA, spatial filtering approaches 
have been proposed. 
In \citep{Friman:2007} the Minimum Energy Combination (MEC) method 
has been proposed while 
in \citep{Lin:2006} the Canonical Correlation Analysis (CCA) method 
was introduced. 
Both methods use sinusoids waves as reference templates and they solve an
optimization problem, based on multi-channel SSVEP data, in order
to obtain optimal spatial filters. 
Finally, extensions of CCA have been proposed 
in \citep{Zhang:2013, Bin:2011, Nakanishi:2015, Nakanishi:2018, Tong:2021, Yuan:2022}, 
extracting the subject-specific and task-related
information from the individual calibration data and reduce
the effect of spontaneous background EEG activities. 

From spatial filtering methods, the task-related component analysis
(TRCA)-based method \citep{Nakanishi:2018} shows great potential since it 
has achieved superior performance among various spatial filtering methods.
The core idea of TRCA is to acquire the spatial filters by 
strengthening the task-related SSVEP components and suppressing the noise.
TRCA-based methods are followed by the target detection step 
where the similarity between the filtered test signal and 
the filtered template is calculated via the correlation coefficient.
All spatial filtering - based methods are based on the basic
(generalized) eigenvalue problem \citep{Oikonomou2020machine, Wong:2020}. 
However, difference between various
approaches can be observed and these difference are reflected to the way 
that the matrices, involved in the eigenvalue problem, are 
constructed \citep{Wong:2020}.
In \citep{Zhang:2018} Correlated Component Analysis 
(CORCA)assumes that the task related component is shared 
among the subjects by adopting 
a transfer learning procedure in the construction of covariance matrices. 
While, in \citep{Liu:2021} a task discriminant component analysis
was applied which involves the construction of within and 
between SSVEP targets covariance matrices.

One defect of TRCA is that it can only deal with limited 
noise components. For other kinds of noise, such as locally 
occurring noises that have the same profile, the TRCA-based 
method is powerless\citep{Wong_2020, Jin:2021}. Additionally, its
performance deteriorate drastically if the calibration 
trials are insufficient. To deal with more general noises and 
the number of trials, we introduce a novel 
adaptive time-domain filter resulting in more reliable similarity 
measurement. By introducing the temporally-based 
filter into the objective function of the TRCA-based method we 
construct a time filter that acts together 
with the spatial filter to suppress more general noises.
Furthermore, the filter adapts to the statistical properties of 
SSVEP trials.

The rest of this paper is organized as follows. In section \ref{sec:Meth}, 
first, we provide a short description of CCA and TRCA concentrating on the mathematical
formulation of them, and then, we describe our approach for SSVEP recognition.
Section \ref{sec:exper}, we describe the SSVEP datasets that are used in our study,
and then we provide details about our experiments and the performance of our method.
Also, a comparison with competing methods is provided. Finally, a short discussion 
and some concluding remarks are are provided in section \ref{sec:Conc}. 

\section{Materials and Methods}\label{sec:Meth}
\subsection{Problem Description}

When a SSVEP experiment is taken place, the subject is seated in front
of a screen where visual stimuli are flashing in different frequencies,
During the experiment raw EEG data are collected in order to
calibrate the overall system. The segmentation of raw EEG data (using event triggers), 
results into a set of trials for each visual stimulus (or class). 
Using these EEG trials the experimenter can calibrated the BCI system 
(for example, by training the classifier). 
Let us assume that the SSVEP dataset is a collection of multi-channel EEG trials $\{ \mathbf{X}_1^{(s)}, \mathbf{X}_2^{(s)}, \cdots, \mathbf{X}_M^{(s)} \}_{s=1}^{N_s}$ for each participant, where $M$ is the number of trials of a SSVEP target, $(s)$ is the index of the SSVEP target, $N_s$ is the number of SSVEP targets (or classes).
Each $\mathbf{X}_m^{(s)}, m=1,\cdots,M, s=1,\cdots,N_s$ is a matrix of $N_{ch} \times N_{t}$, where $N_{ch}$ is the number of channels
and $N_{t}$ the number of samples. Additionally, we assume that 
the multi-channel EEG signals are centralized since in practise the EEG trials are bandpass filtered or detrended. 

\subsection{Canonical Correlation Analysis (CCA)}
Spatial filtering attempts to maximize the SNR between the raw EEG data and the spatial 
filtered version of them. In typical cases, such as bipolar combination or laplacian filtering,
the spatial filters are determined manually. However, this approach does not take into account
any prior knowledge about SSVEPs or any subject-specific information. 
One of the first approach that take into consideration the structure of SSVEPs 
was based on Canonical Correlation Analysis (CCA)\citep{Lin:2006}. 
The CCA is a multivariate statistical method attempting to discover
underlying correlations between two sets of data\citep{Lin:2006,Sun:2014}. 
These two sets of data is assumed to be only a different view (or representation) of the 
same original (hidden) data. More specifically, CCA finds a linear projection 
for each set such that these two set are maximally correlated in the hidden 
(dimensionality-reduced) space.

In the SSVEP problem these two views is the test EEG trial $\mathbf{X}_{m}^{(s)}$ and the reference templates for $s$-th stimulus $\mathbf{Y}_{(f_s)}$, where
$$
\mathbf{Y}_{(f_s)} = \begin{bmatrix}
\sin(2\pi \cdot 1 \cdot f_s t)  \\
\cos(2\pi \cdot 1 \cdot f_s t)   \\
\vdots \\
\sin(2\pi \cdot N_h \cdot f_s t)  \\
\cos(2\pi \cdot N_h \cdot f_s t)
\end{bmatrix}^{\top}
$$
$\mathbf{Y}_{(f_s)} \in \mathbb{R}^{N_t \times 2N_h}$, $f_s$ is the frequency of $s$-th stimulus

Typically, CCA methods maximize the linear correlation between the 
projections $\mathbf{w}_s^T\mathbf{X}_{m}^{(s)}$ 
and $\mathbf{v}_s^T\mathbf{Y}_{f_s}$, where $\mathbf{w}_s \in \mathbb{R}^{N_{ch}}$ and 
$\mathbf{v}_s \in \mathbb{R}^{N_{t}}$. At the end, we solve 
the following optimization problem:
\begin{eqnarray}
\max \rho_s = \max_{\mathbf{w}_s,\mathbf{v}_s} \frac{ \mathbf{w}_s^{\top}\mathbf{X}_{m}^{(s)}\mathbf{Y}_{f_s}^{\top}\mathbf{v}_s }
	{\sqrt{\mathbf{w}_{s}^{\top}\mathbf{X}_{m}^{(s)}(\mathbf{X}_{m}^{(s)})^{\top}\mathbf{w}_s \mathbf{v}_{s}^{\top}\mathbf{Y}_{f_s}\mathbf{Y}_{f_s}^{\top}\mathbf{v}_s } }
\end{eqnarray}
Since $\rho_s$ is invariant to the scaling of $\mathbf{w}_s$ and $\mathbf{v}_s$ 
the above optimization problem can be also formulated as the
following generalized eigenvalue problem:
\begin{eqnarray}
\mathbf{X}_{m}^{(s)} \mathbf{Y}_{f_s}^{\top} (\mathbf{Y}_{f_s}\mathbf{Y}_{f_s}^{\top} )^{-1} \mathbf{Y}_{f_s}(\mathbf{X}_{m}^{(s)})^{\top}\mathbf{w}_s = 
\lambda_s \mathbf{X}_{m}^{(s)}(\mathbf{X}_{m}^{(s)})^{\top}\mathbf{w}_s
\end{eqnarray}
where $\lambda_s$ is the eigenvalue corresponding to the eigenvector $\mathbf{w}_s$.

In order to find the stimulus of test EEG trial $\mathbf{X}_{m}^{(s)}$, that 
the subject intends to select, we find $\rho_s$ for all
available stimuli and the stimulus-target, $c$, is then identified by finding the index of the
maximum feature among $N_s$ features:
$c = \arg \max_{s} \{\rho_s\}$. It must be observed here that there is no need 
for training (or calibration) since the templates $\mathbf{Y}_{f_s}$ are artificially
generated. 

\subsection{Task Related Component Analysis (TRCA)}
Task-related component analysis (TRCA) enhances reproducibility of SSVEPs across multiple trials and the intuition of TRCA is to maximize the reproducibility of 
SSVEP target-related components after spatial filtering.
More specifically, the TRCA method find the spatial filters $\mathbf{w}_s$ by solving a generalized
eigenvalue problem which is described by the following equation:
\begin{eqnarray}
\max_{\mathbf{w}_s} \frac{ \mathbf{w}_s^{\top}AA^{\top}\mathbf{w}_s } {\mathbf{w}_s^{\top}BB^{\top}\mathbf{w}_s }
\end{eqnarray}
where $A = \frac{1}{M}\sum_{m=1}^{M} \mathbf{X}_{m}^{(s)}$, and $B$ is a concatenated matrix contains
all trials of $s$-th stimulus, 
$B = [ \mathbf{X}_1^{(s)},  \mathbf{X}_2^{(s)}, \cdots \mathbf{X}_M^{(s)} ]$. 

In order to find the target of the test trial, $\mathbf{X}_{test}$ 
we apply the following discriminant function:
\begin{equation} 
c = \arg \max_{s} \{corr(\mathbf{w}_s^{\top}\mathbf{X}_{test}, \mathbf{w}_s^{\top}A) \}
\end{equation}
where $corr(\cdot,\cdot)$ denotes the Pearson's correlation coefficient.

\subsection{Adaptive Task-Related Component Analysis (adTRCA)}
In our work we propose a new generalized
eigenvalue problem for SSVEP detection which is described by the following equation:
\begin{eqnarray}
\label{eq:adGen}
\max_{\mathbf{w}_s} \frac{ \mathbf{w}_s^{\top}A C A^{\top}\mathbf{w}_s } {\mathbf{w}_s^{\top}B D B^{\top}\mathbf{w}_s }
\end{eqnarray}
where $C$ and $D$ are "filtering" matrix that acts on the time dimension of trials.
The matrices $C$ and $D$ can be defined using various approaches and 
their goal is to remove noise in time domain.

In our study we make some critical assumptions about the generation model 
of SSVEP responses, which affect the data analysis procedure. 
More specifically, SSVEP responses contains strong sinusoids components \citep{Lin:2006},
hence the SSVEP signal in each channel is modeled as a linear 
combination of sinusoids described the following matrix:
\begin{equation} 
\boldsymbol{\Phi} = [ \mathbf{Y}_{(f_1)} \mathbf{Y}_{(f_2)} \cdots 
\mathbf{Y}_{(f_{N_s})} ] \in \mathbb{R}^{N_t \times (2 N_s N_h)}. \nonumber 
\end{equation}
Additionally, SSVEP responses belonging to the same visual stimulus share 
common components. From the above we can observe that the generation
of SSVEP responses can be modeled as multiple regression tasks that share
common information. 


EEG trials from the $s$-th stimulus are collected in matrix 
$\mathbb{B} = [ {\mathbf{X}_{1}^{(s)}}^{\top},  {\mathbf{X}_{2}^{(s)}}^{\top}, \cdots {\mathbf{X}_{M}^{(s)}}^{\top} ]^{\top} $,
$\mathbb{B} \in \mathbb{R}^{N_t \times (N_{ch} N_s)}$, where 
each column of $\mathbb{B}$ contains the data from one channel 
or each column of $\mathbb{B}$ contains the data from one task. Hence we have
$\boldsymbol{y}_i \in \mathbb{R}^{N_t \times 1} , i=1,\cdots, N_{ch} N_s$ tasks (the i-th column of $\mathbb{B}$).
Each learning task can be described by the following linear regression model:
\begin{equation}
\label{Eq:RegM}
    \boldsymbol{y}_i = \boldsymbol{\Phi}\boldsymbol{w}_i + \boldsymbol{e}_i
\end{equation}
where 
$\boldsymbol{w}_i$ $2 N_s N_h \times 1$ vector of weights (or parameters), and,
$\boldsymbol{e}_i$ $N_t \times 1$ vector of noise coming from a zero mean
Gaussian random variable with unknown precision (inverse variance) $a_0$.
We can observe that each of the mapping yields a corresponding regression task, 
and performing multiple such learning tasks has been referred to as multitask learning \citep{Ji:2009}, 
which aims at sharing information effectively among multiple related tasks.
In a more abstract view of our problem we can see that each learning task is 
a linear regression problem, and sinusoids components from one regression task affect the fitting procedure of another regression task. 




The likelihood function for parameters $\boldsymbol{w}_i$ and $a_0$ is given by:
\begin{equation}
    p(\boldsymbol{y}_i | \boldsymbol{w}_i,a_0) = (2\pi a_0)^{-\frac{N_t}{2}} 
    \exp \Big\{ -\frac{a_0}{2}  \|\boldsymbol{y}_i - \boldsymbol{\Phi}\boldsymbol{w}_i\|^{2}_{2}  \Big\}
\end{equation}
The parameters of a regression task, $\boldsymbol{w}_i$, are assumed to be drawn 
from a product of zero-mean Gaussian distributions that are shared by all tasks. 
Letting $w_{i,j}$ be the $j$-th parameters for $i$-th task then we have:
\begin{equation}
    p(\boldsymbol{w}_i|\boldsymbol{a}) = \prod_{j=1}^{2 N_s N_h} {\cal N} (w_{i,j} | 0, a^{-1}_i )
\end{equation}
where the hyperparameters $\boldsymbol{a} = \{a_j\}_{j=1,2,\cdots,2 N_s N_h}$ are 
shared 
among $N_{ch} N_s$ regression tasks, hence, data from all regression tasks 
contribute to learning 
these hyperparameters.
To promote sparsity over parameters, we place Gamma priors over hyperparameters
$\boldsymbol{a}$\citep{Tipping:2001,Ji:2009}. 
Also, the same type of prior is placed over noise precision $a_0$.

\begin{align}
    p(a_0|\alpha,\beta) & = Ga(a_0|\alpha,\beta) \nonumber \\
    & = \frac{\beta^{\alpha}}{\Gamma(\alpha)} a_{0}^{\alpha - 1} \exp\Big\{  -\beta a_0\Big\}
\end{align}

\begin{equation}
    p(\boldsymbol{a} | c,d) = \prod_{j=1}^{2 N_s N_h} Ga(a_j|c,d)
\end{equation}
In addition, we can observed here, that noise properties are shared among different tasks
(i.e. the noise vectors in Eq. (\ref{Eq:RegM}) are drawn 
from the same Gaussian distribution).
Finally, it must be noted that we have an hierarchical model, and these types
of models are natural to be "dealt" within the bayesian framework.

Given hyperparameters $\boldsymbol{a}$ and noise precision $a_0$, 
we can apply Bayes theorem
to find the posterior distribution over $\boldsymbol{w}_i$, which is a 
Gaussian distribution:

\begin{align}
    p(\boldsymbol{w}_i|\boldsymbol{y}_i,\boldsymbol{a},a_0) &= 
    \frac{p(\boldsymbol{y}_i | \boldsymbol{w}_i,a_0)p(\boldsymbol{w}_i)|\boldsymbol{a})}
    {\int p(\boldsymbol{y}_i | \boldsymbol{w}_i,a_0)p(\boldsymbol{w}_i)|\boldsymbol{a}) d\boldsymbol{w}_i }
    \nonumber \\
    &={\cal N}\Big(\boldsymbol{w}_i \Big| \boldsymbol{\mu}_i,\boldsymbol{\Sigma}_i\Big)
\end{align}

where 

\begin{align}
\label{Eq:mu}
    \boldsymbol{\mu}_i & = a_0 \boldsymbol{\Sigma}_i \boldsymbol{\Phi}^{T}\boldsymbol{y}_i \\
\label{Eq:Sigma}    
    \boldsymbol{\Sigma}_i & = \Big( a_0 \boldsymbol{\Phi}^{T}\boldsymbol{\Phi} + \boldsymbol{A}\Big)^{-1} 
\end{align}
and $\boldsymbol{A} = diag(a_1,a_2,\cdots,a_M)$. 

In order to find hyperparameters $\boldsymbol{a}$ and promote sparsity in parameters, 
the type-II Maximum Likelihood procedure is adopted \citep{Tipping:2001,MacKay_ard}, 
where the objective is to maximize
the marginal likelihood (or its logarithm). Also, a similar procedure is followed for
the noise precision. The marginal likelihood ${\cal L}( \boldsymbol{a},a_0 )$ is given by:

\begin{align} 
 & {\cal L}( \boldsymbol{a},a_0 )  = \sum_{i=1}^{L} \log
    \int p(\boldsymbol{y}_i | \boldsymbol{w}_i,a_0) p(\boldsymbol{w}_i|\boldsymbol{a}) d\boldsymbol{w}_i 
    \nonumber \\
    &= -\frac{1}{2} \sum_{i=1}^{L} \Big( N_i \log(2\pi) + \log |\boldsymbol{C}_i| + 
    \boldsymbol{y}_{i}^{T} \boldsymbol{C}_{i}^{-1} \boldsymbol{y}_{i} \Big) 
\end{align}
where $\boldsymbol{C}_{i} = a_{0}^{-1}\boldsymbol{I} + 
\boldsymbol{\Phi}\boldsymbol{A}\boldsymbol{\Phi}^{T} $

Differentiating ${\cal L}( \boldsymbol{a},a_0 )$ with respect to $\boldsymbol{a}$ 
and $a_0$ and setting the results into zero \citep{Tipping:2001,MacKay_ard,Ji:2009} 
(after some algebraic manipulations)
we obtain:
\begin{align}
\label{Eq:a}
    a_{j}^{(new)} & = \frac{(N_{ch} N_s) - a_j\sum_{i=1}^{(N_{ch} N_s)} \Sigma_{i,(j,j)}}{\sum_{i=1}^{(N_{ch} N_s)}\mu_{i,j}},
    j=1,2,\cdots,2 N_s N_h \\
\label{Eq:a0}    
    a_{0}^{(new)} & = \frac{\sum_{i=1}{N_{ch} N_s}\Big( N_t - 2 N_s N_h + \sum_{j=1}^{2 N_s N_h} a_{j} \Sigma_{i,(j,j)} \Big)}    {\sum_{i=1}^{N_{ch} N_s} \|\boldsymbol{y}_i - \boldsymbol{\Phi}_i\boldsymbol{\mu}_i\|^{2}_{2}}
\end{align}
where $\mu_{i,j}$ the $j$-th element of $\boldsymbol{\mu}_i$ and
$\Sigma_{i,(j,j)}$ the $j$-th diagonal element of covariance matrix $\boldsymbol{\Sigma}_i$.
The above analysis suggests an iterative algorithm that iterates between Eqs. 
(\ref{Eq:mu}), (\ref{Eq:Sigma}), (\ref{Eq:a}) and (\ref{Eq:a0}), until
a convergence criterion is satisfied. Also, the same algorithm can be derived by 
adopting the EM framework and treating parameters $\boldsymbol{w}_i$ 
as hidden variables\citep{Tipping:2001}. Finally, based on the above bayesian formulation,
we can derive a fast version of the above algorithm. 
The fast version provides an elegant treatment of feature vectors by constructing 
adaptively the matrix $\boldsymbol{\Phi}$ through three basic operators:
addition, deletion and re-estimation. More information on this subject
can be found in \citep{Tipping:2001,Ji:2009}.

Now, SSVEP components in each task can be represented as:
$$
\boldsymbol{\hat{y}}_i = \boldsymbol{\Phi}\boldsymbol{\mu}_i, 
i=1,\cdots, N_{ch} N_s
$$
rearranging filtered EEG signals, $\boldsymbol{\hat{y}}_i$, each filtered EEG trial 
is represented as:
$\mathbf{X^f}_m^{(s)} = \mathbf{X}_m^{(s)}a_0\boldsymbol{\Phi}\boldsymbol{\Sigma}_i\boldsymbol{\Phi}^{\top}$. 
Due to filtered trials we find the spatial filters $\mathbf{w}_s$ by solving the
following generalized eigenvalue problem:
\begin{eqnarray}
\label{eq:filterTRCA}
\max_{\mathbf{w}_s} \frac{ \mathbf{w}_s^{\top}A_fA_f^{\top}\mathbf{w}_s } {\mathbf{w}_s^{\top}B_fB_f^{\top}\mathbf{w}_s }
\end{eqnarray}
where $A_f = \frac{1}{M}\sum_{m=1}^{M} \mathbf{X^f}_{m}^{(s)}$, and $B_f$ is a concatenated matrix contains
all trials of $s$-th stimulus, 
$B_f = [ \mathbf{X^f}_1^{(s)},  \mathbf{X^f}_2^{(s)}, \cdots \mathbf{X^f}_M^{(s)} ]$. 
The above generalized eigenvalue problem can be connected by that of Eq. \ref{eq:adGen}.
After some algebraic manipulations Eq. \ref{eq:filterTRCA} can be written as:
\begin{eqnarray}
\max_{\mathbf{w}_s} \frac{ \mathbf{w}_s^{\top}A C A^{\top}\mathbf{w}_s } {\mathbf{w}_s^{\top}B D B^{\top}\mathbf{w}_s }
\end{eqnarray}
where $C = (a_0\boldsymbol{\Phi}\boldsymbol{\Sigma}_i\boldsymbol{\Phi}^{\top})
(a_0\boldsymbol{\Phi}\boldsymbol{\Sigma}_i\boldsymbol{\Phi}^{\top})^{\top}$ and
$
D = 
\begin{bmatrix}
C & \cdots & 0 \\
\vdots & \ddots & \vdots \\
0 & \cdots & C \\
\end{bmatrix}.
$ We can observe an interesting connection between the proposed
method and the TRCA. When the $C=\mathbf{I}$, where $\mathbf{I}$ is
the unitary matrix, the proposed approach degraded to the TRCA
method. We see that the TRCA method is a limiting case of the 
proposed method. In addition, we can observe that matrices $C$ and 
$D$ acts on the time dimension of the EEG trials, hence time samples
are differently treated according to the time dimension rather
than equally weighted. Also, we can observe that filters, represented
by matrix $C$, are adapted to the statistical properties of EEG trials.  
Finally, after finding the spatial filters, to find the target of the test trial, $\mathbf{X}^{f}_{test}$ 
we apply the following discriminant function:
\begin{equation} 
\label{Eq:DiscrRule1}
c = \arg \max_{s} \{corr(\mathbf{w}_s^{\top}\mathbf{X}^{f}_{test}, \mathbf{w}_s^{\top}A_f) \}
\end{equation}

\subsection{Ensemble case}\label{sec:ensemble}
According to the previous discriminant rule, described by Eq. \ref{Eq:DiscrRule1},
we can observe that to calculate the similarity of the test trial to the 
stimulus $s$ first we apply spatial filter $\mathbf{w}_s$. However, since
we have enough calibration trials we can obtained $N_s$ spatial filters for
each stimulus \citep{Nakanishi:2018,Zhang:2018}. Hence,
we can extend our method using an ensemble approach, similar to 
\citep{Nakanishi:2018,Zhang:2018}, where all spatial filters
for stimulus $s$ are concatenated to create an ensemble spatial filter 
$\mathbf{W}_s \in \mathbb{R}^{N_{ch} \times N_{s}}$.
Now, in order to find the target of the test trial, $\mathbf{X}^{f}_{test}$ 
we apply the following discriminant function:
\begin{equation} 
c = \arg \max_{s} \{corr(\mathbf{W}_s^{\top}\mathbf{X}^{f}_{test}, \mathbf{W}_s^{\top}A_f) \},
\end{equation}
where, now, the function $corr(\cdot)$ depicts correlation between matrices.

\section{Results}\label{sec:exper}
In our study two, publicly available, SSVEP datasets are used to evaluate our method.
We call these datasets, the \textit{Speller} dataset and the 
\textit{EPOC dataset}, and we provide a description of them in the following paragraphs.

\textit{Speller} dataset \citep{YWang:2017}: 
The \textit{Speller} dataset \citep{YWang:2017} contains 
40-target visual stimuli were presented on a 23.6-in LCD monitor. Thirty five 
healthy subjects (with normal or corrected-to-normal vision) were recruited 
for the SSVEP experiment. Furthrmore, eight of them had experience 
with using a SSVEP-based BCI speller. 
EEG signals were recorded with 64 channels according to an extended 10--20 system.
In our experiments we have used electrodes from the occipital and 
parietal-occipital areas ($P_z$, $PO_5$, $PO_3$, $PO_z$, $PO_4$, $PO_6$, $O_1$, $O_z$, and $O_2$) . 
During the experiment, each subject completes 6 blocks, where in each block
the subject was looking at one of the visual stimuli, indicated by the stimulus program,
in a randomg order for 5s and he/she complete 40 trials corresponding to all 40 targets. 
Using event triggers, EEG trials were extracted and down-sampled to 250Hz. Futhermore,
there trials have been band-pass filterd to 7-90Hz with an infinite impulse response (IIR) 
filter using the \textit{filtfilt} function from MATLAB. 
Additionally, a delay of 140ms in the visual system was considered 
in our experiments\citep{YWang:2017}. 

\textit{EPOC} dataset~\citep{Oikonomou:2016}: 
EEG trials, from 11 subjects executing a SSVEP-based
experimental protocol, were acquired. The frequencies of visual 
stimulation were: 
6.66Hz, 7.50Hz, 8.57Hz, 10.00Hz and 12.00Hz. 
EEG trials were recorded using the Emotiv Epoc device, with 14 wireless
channels and a sampling rate of 128 Hz.
Each subject was asked to gaze at one of the visual stimuli, indicated by the stimulus program, 
in a random order, and complete 20 trials for each of the five targets. 
The EEG data have been band-pass filtered 
from 5Hz to 45Hz. More information about this dataset can be found in 
\citep{Oikonomou:2016} and https://physionet.org/content/mssvepdb/1.0.0/.
The visual latency is an important factor in a SSVEP BCI system. 
An accurate estimation of the visual latency ensures that 
the extracted data epochs only contain SSVEP responses to the stimulation. 
In the \textit{Speller} dataset the visual latency is defined using the classification 
accuracy\citep{YWang:2017}, while, in the \textit{EPOC dataset} an alignment procedure is used to extract data epochs containing the SSVEP responses.

\subsection{Performance metrics}
To calculate the performance metrics we adopt a leave-one-out cross-validation scheme where we 
use $B-1$ blocks for training and 1 block for testing, similar to
\citep{Nakanishi:2015,Wang:2016,YWang:2017,Oikonomou:2019jbhi}. 
The performance of the examined algorithms was calculated by using the following metrics:

\textbf{Classification Accuracy:}
Classification Accuracy is the ratio between the number of correctly classified trials 
to the total number of trials.

\textbf{Information Transfer Rate (ITR):}
When a BCI system is designed, we are interested, besides the accuracy, to 
the number of classes offered, as well as, to the time required for the classification process.
These two factors are particularly important in the design of BCI systems since they determine the amount of information that can be transmitted. Hence, we use an additional metric called
Information Transfer Rate (ITR). ITR is used to quantify the rate of information transmitted by the BCI system and is measured by the following equation: 
\begin{equation}
\label{eq:ITR}
ITR = \Bigg( \log_2 K + P\cdot\log_2 P + (1-P)\log_2 \Big[\frac{1-P}{K-1}\Big] \Bigg)\cdot \Bigg( \frac{60}{T}\Bigg)
\end{equation}
with $K$ being the number of classes, $P$ the classification accuracy, and $T$ the time in seconds required for the classification process to complete.

We compare the proposed spatial filtering method with two spatial 
filtering approaches: the CCA\citep{Lin:2006} and the TRCA\citep{Nakanishi:2018},
and with two ML methods: the MLR approach \citep{Wang:2016} and the Graph-based 
Sparse Representations Classification (MLR-SRC) \citep{Oikonomou:2021}.
We have calculate the above metrics with respect the time window of the 
test EEG trial, the number of EEG channels and the number of training trials.

\subsection{Performance Comparison versus the Time Window Length}
\begin{figure*}
	\begin{center}
    \begingroup 
		\renewcommand\tabcolsep{1pt}
		\begin{tabular}[c]{cc}
			\includegraphics[width=8cm]{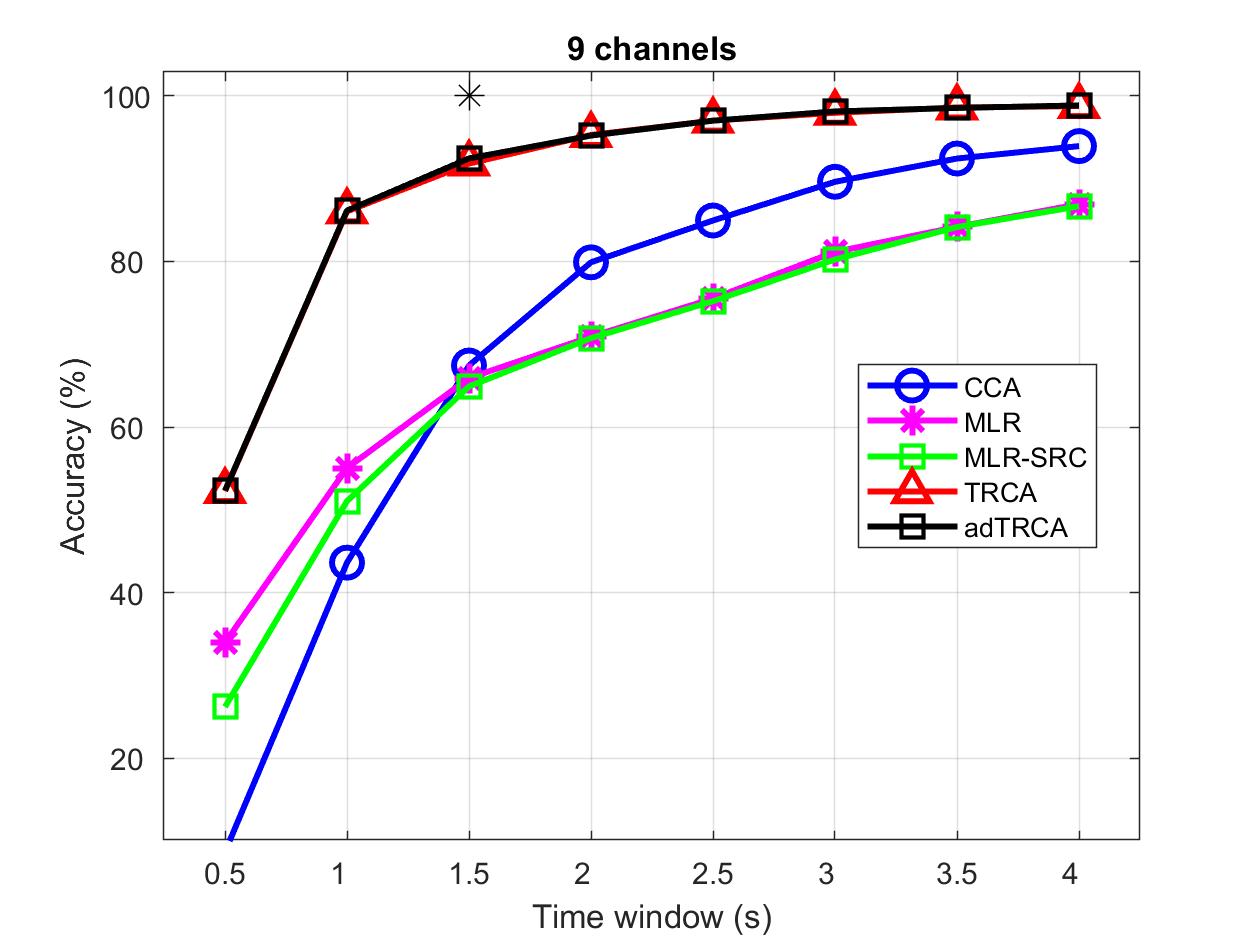} &
            \includegraphics[width=8cm]{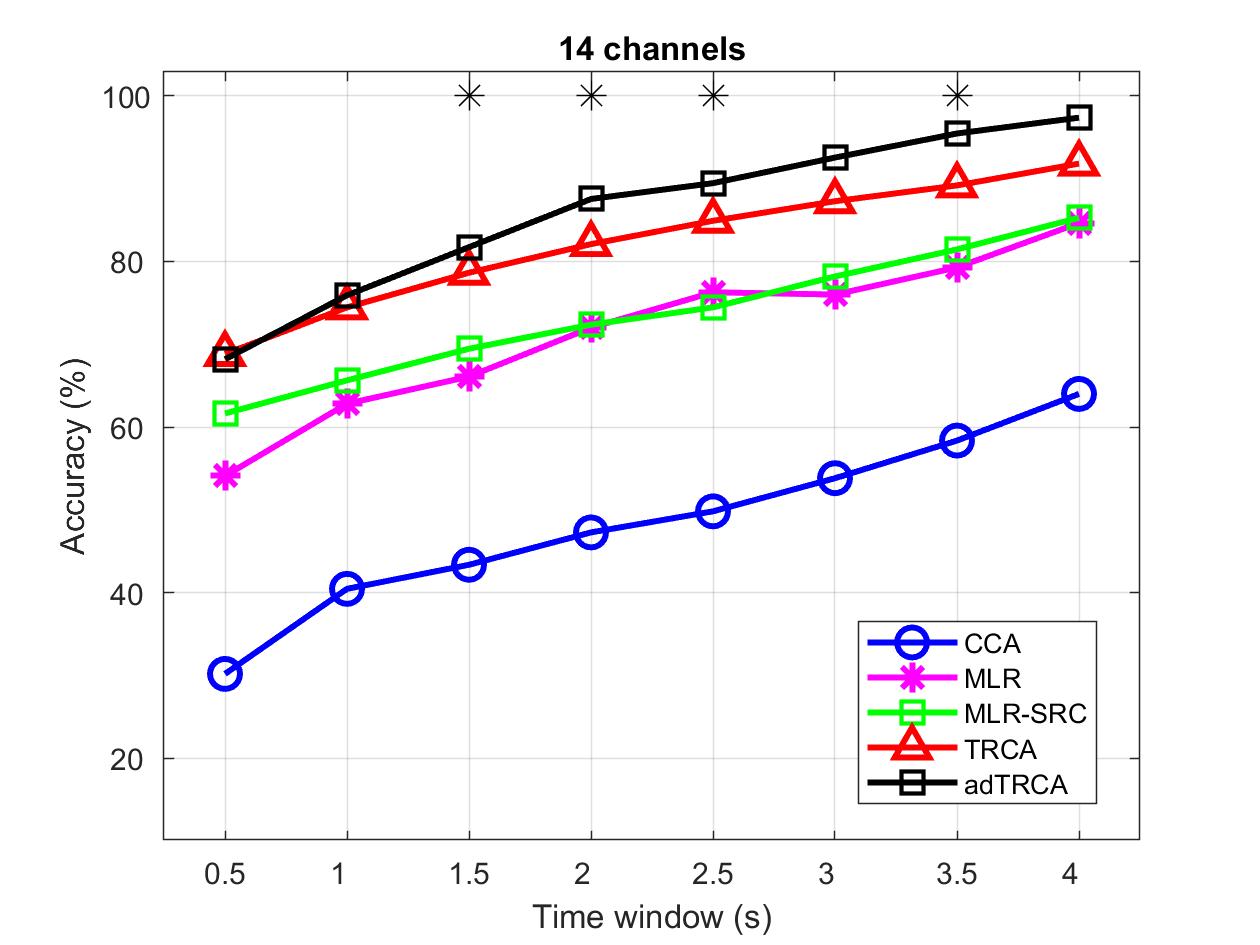} \\
			(a) & (b) \\
		\end{tabular}
    \endgroup
	\end{center}
	\caption{Average Classification over all subjects (a) for the Speller dataset and (b) the EPOC dataset 14 using the basic configuration with respect to the EEG channels.  In both cases, the time window ranges from 0.5s to 4s (0.5s interval). *  indicates statistically 
	significant difference between the TRCA 
	and adTRCA methods, using paired sample t-test for Speller dataset and Wilcoxon signed rank test for the EPOC dataset ($p < 0.05$).}
	\label{fig:resBasic9}
\end{figure*}

In the first series of experiments we follow a typical analysis of SSVEP datasets
by examine the performance of methods with respect to the time window.
More specifically, the performance of methods are evaluated for variable time
from 0.5sec to 4sec with step of 0.5sec. In addition for the Speller dataset
we use the 9 channels from occipital and parietal - occipital areas 
(Pz, PO5, PO3, POz, PO4, PO6, O1, Oz, O2), while,
for the EPOC dataset we use all available 14 channels, covering the entire brain.
In Fig. \ref{fig:resBasic9} we provide the obtained results for all
comparative methods for the two datasets using the basic channel configuration.
We can observed the adTRCA and TRCA methods provides better results from
all other methods in both datasets. Also, we can observed that between 
the adTRCA and TRCA, the adTRCA method has marginally better detection accuracy
in the \textit{Speller} dataset, and significant better detection accuracy in the
\textit{EPOC} dataset. Similar conclusions can be drawn with respect to the ITR
(see Tables \ref{tbl:ItrLargeDataset9chan} and \ref{tbl:ItrEpocDataset14chan}).
We can observed that for most TWs, the adTRCA methods provides the best ITR 
among all methods. However, we must point out, that for the \textit{Speller}
dataset the best ITR is achieved at TW=0.5s from the adTRCA method, while,
for the \textit{EPOC} dataset, the best ITR value is achieved at TWs=0.5s from the
TRCA method. 


\begin{table}[h]
\begin{center}
\caption{ITR on Speller dataset - 9 channels}
\begin{tabular}{|c|c|c|c|c|c|} \hline
TW	& CCA 	    &MLR	& MLR-SRC	& TRCA	& adTRCA \\ \hline
0.5 &  27.3581 & 141.5439  & 98.0103 & 270.9573 & \textbf{271.3097} \\ \hline
1   &  101.8150 & 137.3785 & 123.1899 & 260.0963 & \textbf{260.3216} \\ \hline
1.5 &  122.8637 & 118.8840 & 116.4590 & 189.4191 & \textbf{190.8013} \\ \hline
2   &   116.7986 &  98.5886 &  98.1738 & \textbf{149.0613} & 148.8653 \\ \hline
2.5 &  101.4344  & 85.9023  & 85.3361  & \textbf{122.2336} & 122.1442 \\ \hline
3   &    90.6986 &  79.0954 &  77.8773 & 103.1242 & \textbf{103.3462} \\ \hline
3.5 &   81.1131  & 71.3472  & 71.3823  & \textbf{89.2204}  & 89.0756\\ \hline
4   &   72.5963  & 65.1814  & 65.0779  & 78.2199  & \textbf{78.3041} \\ \hline
\end{tabular}
\label{tbl:ItrLargeDataset9chan}
\end{center}
\end{table}


\begin{table}[h]
\begin{center}
\caption{ITR on EPOC dataset - 14 channels}
\begin{tabular}{|c|c|c|c|c|c|} \hline
TW	& CCA 	    & MLR	& MLR-SRC	& TRCA	& adTRCA \\ \hline
0.5 &  13.7550 &  61.4288 &  81.6831 & \textbf{113.1879} & 109.7278   \\ \hline
1   &  16.4944 &  45.1581 &  49.0298 &  67.8476 &  \textbf{70.7059}  \\ \hline
1.5 &  13.3129 &  35.7231 &  37.9279 &  52.7534 &  \textbf{55.7158} \\ \hline
2   &  13.6286 &  31.7665 &  31.8755 &  44.0878 &  \textbf{48.9809}  \\ \hline
2.5 &  12.4521 &  29.7089 &  28.1211 &  38.7167 &  \textbf{42.2460}  \\ \hline
3   &  12.0830 &  25.1981 &  26.2000 &  35.0314 &  \textbf{38.0493} \\ \hline
3.5 &  12.1995 &  23.8428 &  24.4427 &  31.7552 &  \textbf{34.8263}   \\ \hline
4   &  12.6687 &  23.6196 &  23.6422 &  29.8733 &  \textbf{32.1765}  \\ \hline
\end{tabular}
\label{tbl:ItrEpocDataset14chan}
\end{center}
\end{table}

\subsection{Performance Comparison using the minimal number of EEG channels}

In the second series of experiments we have used the minimal number of EEG channels
to evaluate the performance of methods. These EEG channels
covering the occipital area depending of the EEG device that had been used 
in each dataset. More specifically, for the Speller dataset O1, O2 and Oz EEG channels 
are used, while, for the EPOC dataset, we use O1 and O2 channels. 
This study could correspond to cases where we are not able to use high density
EEG devices such as in BCI applications outside a controlled environment.

\begin{figure*}
	\begin{center}
    \begingroup 
		\renewcommand\tabcolsep{1pt}
		\begin{tabular}[c]{cc}
			\includegraphics[width=8cm]{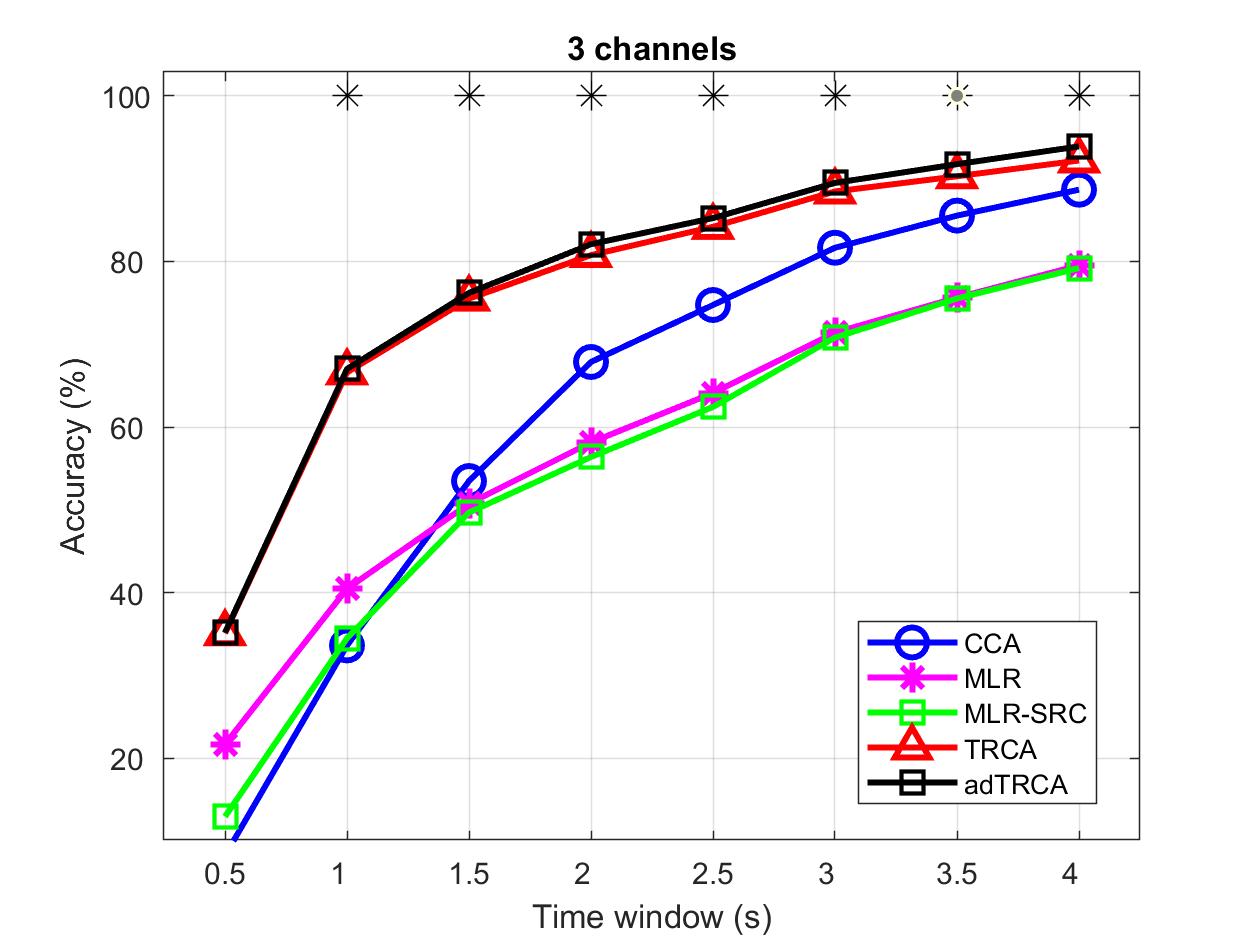} &
            \includegraphics[width=8cm]{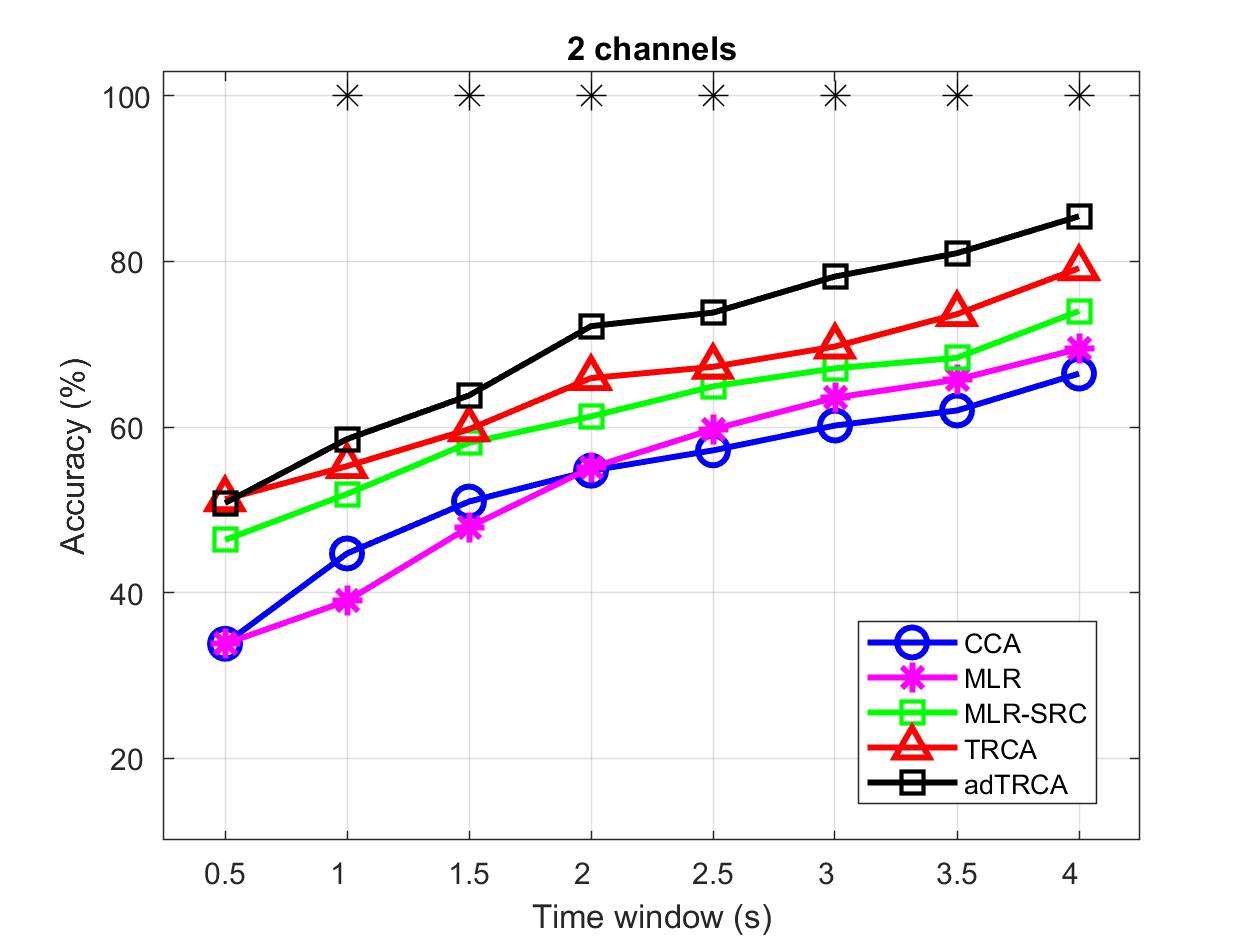} \\
			(a) & (b) \\
		\end{tabular}
    \endgroup
	\end{center}
	\caption{Average Classification over all subjects (a) for the Speller dataset (b) for the EPOC dataset using the EEG channels covering the occipital areas.  In both cases, the time window ranges from 0.5s to 4s (0.5s interval). 
	*  indicates statistically significant difference 
	between the TRCA  and adTRCA methods, using paired sample t-test 
	for Speller dataset and Wilcoxon signed rank test for 
	the EPOC dataset ($p < 0.05$).}
	\label{fig:resBasic3}
\end{figure*}

In Fig. \ref{fig:resBasic3} we provide the obtained results for all
comparative methods for the two datasets using the minimal channels' configuration.
We can observed the adTRCA and TRCA methods provides better results from
all other methods in both datasets. Also, we can observed that between 
the adTRCA and TRCA, the adTRCA method has better significant better detection accuracy
in both datasets. 
In Tables \ref{tbl:ItrLargeDataset3chan} and \ref{tbl:ItrEpocDataset2chan},
we provide the results with respect to the ITR measure for all methods.
We can observed that for all TWs, besides ones, the adTRCA methods provides the best ITR 
among all methods. However, while for the \textit{Speller}
dataset the best ITR is achieved at TW=0.5s from the adTRCA method,
for the \textit{EPOC} dataset, the best ITR value is achieved at the same TW from the
TRCA method. 

\begin{table}[h]
\begin{center}
\caption{ITR on Speller dataset - 3 channels}
\begin{tabular}{|c|c|c|c|c|c|} \hline
TW	& CCA 	    & MLR	& MLR-SRC	& TRCA	& adTRCA \\ \hline
0.5 &  26.6513  & 82.8903 &  39.9318 & 162.8981 & \textbf{163.3228}  \\ \hline
1   &  71.6644  & 93.4390 &  74.2354 & 189.5521 & \textbf{190.6936}  \\ \hline
1.5 &  89.8807  & 85.1997 &  82.2758 & 148.7973 & \textbf{150.0271} \\ \hline
2   &  93.5801  & 76.2840 &  73.1798 & 121.8152 & \textbf{123.6819}   \\ \hline
2.5 &  85.4184  & 69.7438 &  67.2738 & 102.4240 & \textbf{103.7595}  \\ \hline
3   &  80.0041  & 67.1743 &  66.4061 &  90.6681 &  \textbf{91.6374} \\ \hline
3.5 &  73.1382  & 62.4800 &  62.2848 &  79.6201 &  \textbf{81.0671}   \\ \hline
4   &  67.1804  & 58.2207 &  57.9068 &  71.3779 &  \textbf{72.9298} \\ \hline
\end{tabular}
\label{tbl:ItrLargeDataset3chan}
\end{center}
\end{table}


\begin{table}[h]
\begin{center}
\caption{ITR on EPOC dataset - 2 channels}
\begin{tabular}{|c|c|c|c|c|c|} \hline
TW	& CCA 	    & MLR	& MLR-SRC	& TRCA	& adTRCA \\ \hline
0.5 &  20.4758 &  17.7383 &  41.2960 &  \textbf{55.2841} &  54.7309   \\ \hline
1   &  20.8587 &  13.2992 &  29.1648 &  34.8970 &  \textbf{38.5391}  \\ \hline
1.5 &  20.2541 &  16.2247 &  26.1393 &  29.4451 &  \textbf{31.4028} \\ \hline
2   &  18.8570 &  16.9955 &  21.9714 &  27.6870 &  \textbf{31.5392}  \\ \hline
2.5 &  16.7189 &  17.8696 &  20.5254 &  23.4503 &  \textbf{27.1407}  \\ \hline
3   &  15.7776 &  16.6921 &  18.1873 &  20.4231 &  \textbf{25.5093} \\ \hline
3.5 &  14.5984 &  15.0986 &  16.8087 &  19.7346 &  \textbf{23.7536}   \\ \hline
4   &  14.6751 &  15.3353 &  17.0919 &  19.9582 &  \textbf{23.4404}  \\ \hline
\end{tabular}
\label{tbl:ItrEpocDataset2chan}
\end{center}
\end{table}

\subsection{Performance Comparison versus the Number of training trials}

In the last series of experiments, we investigate how the performance of TRCA and adTRCA
are affected by the number of training trials when the time window is 1sec. 
The obtain results are provided in Tables \ref{tbl:AccLargeDataset2chanTr} 
and  \ref{tbl:AccEpocDataset2chanTr}. The proposed method clearly has better accuracy
from TRCA. Additionally, the proposed scheme achieves the best performance among them
and the margin provided by it is more distinct when a smaller
number of training blocks are utilized. Especially, in the case Speller dataset, 
when only three training blocks are utilized, the proposed scheme achieves a
classification accuracy of 63\%, while the TRCA method achieve an accuracy of
61\%. Furthermore, in the case of EPOC dataset, when three training blocks are used
the proposed scheme achieves a classification accuracy of 33\%, while 
the TRCA method achieve an accuracy of 22\% (slighlty above the random guess).

\begin{table}[h]
\begin{center}
\caption{Classification accuracy (\%) on Speller dataset with respect to the number of training trials - 9 channels}
\begin{tabular}{|c|c|c|} \hline
Num	& TRCA	& adTRCA \\ \hline
3   & 61.5714 &  \textbf{63.0238} \\ \hline
4   & 82.9429 &  \textbf{83.3000} \\ \hline
5   & 82.9429 &  \textbf{83.3000} \\ \hline
6   & 86.0476 &  \textbf{86.1429} \\ \hline
\end{tabular}
\label{tbl:AccLargeDataset2chanTr}
\end{center}
\end{table}

\begin{table}[h]
\begin{center}
\caption{Classification accuracy (\%) on EPOC dataset with respect to the number of training trials - 2 channels}
\begin{tabular}{|c|c|c|} \hline
Num	& TRCA	& adTRCA \\ \hline
3  &  22.4242 &  \textbf{33.3333} \\ \hline
7  &  36.8831  & \textbf{39.4805} \\ \hline
10  &  37.6364 &  \textbf{39.8182} \\ \hline
15  &  42.7879 &  \textbf{42.7879} \\ \hline
20  &  54.5455 &  \textbf{55.2727} \\ \hline
\end{tabular}
\label{tbl:AccEpocDataset2chanTr}
\end{center}
\end{table}


\subsection{Ensemble case - Experiments}
\begin{figure*}
	\begin{center}
    \begingroup 
		\renewcommand\tabcolsep{1pt}
		\begin{tabular}[c]{cc}
			\includegraphics[width=8cm]{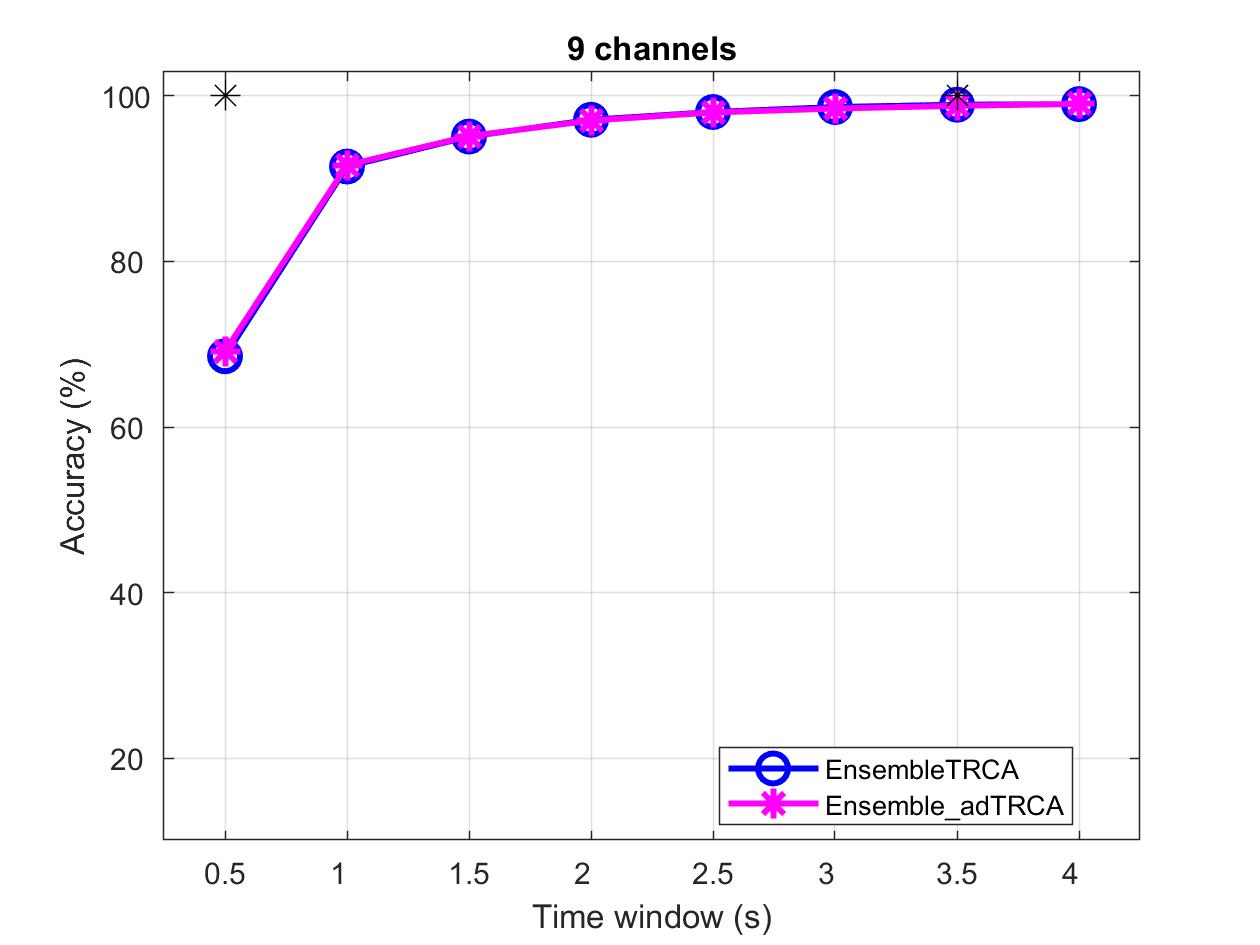} &
            \includegraphics[width=8cm]{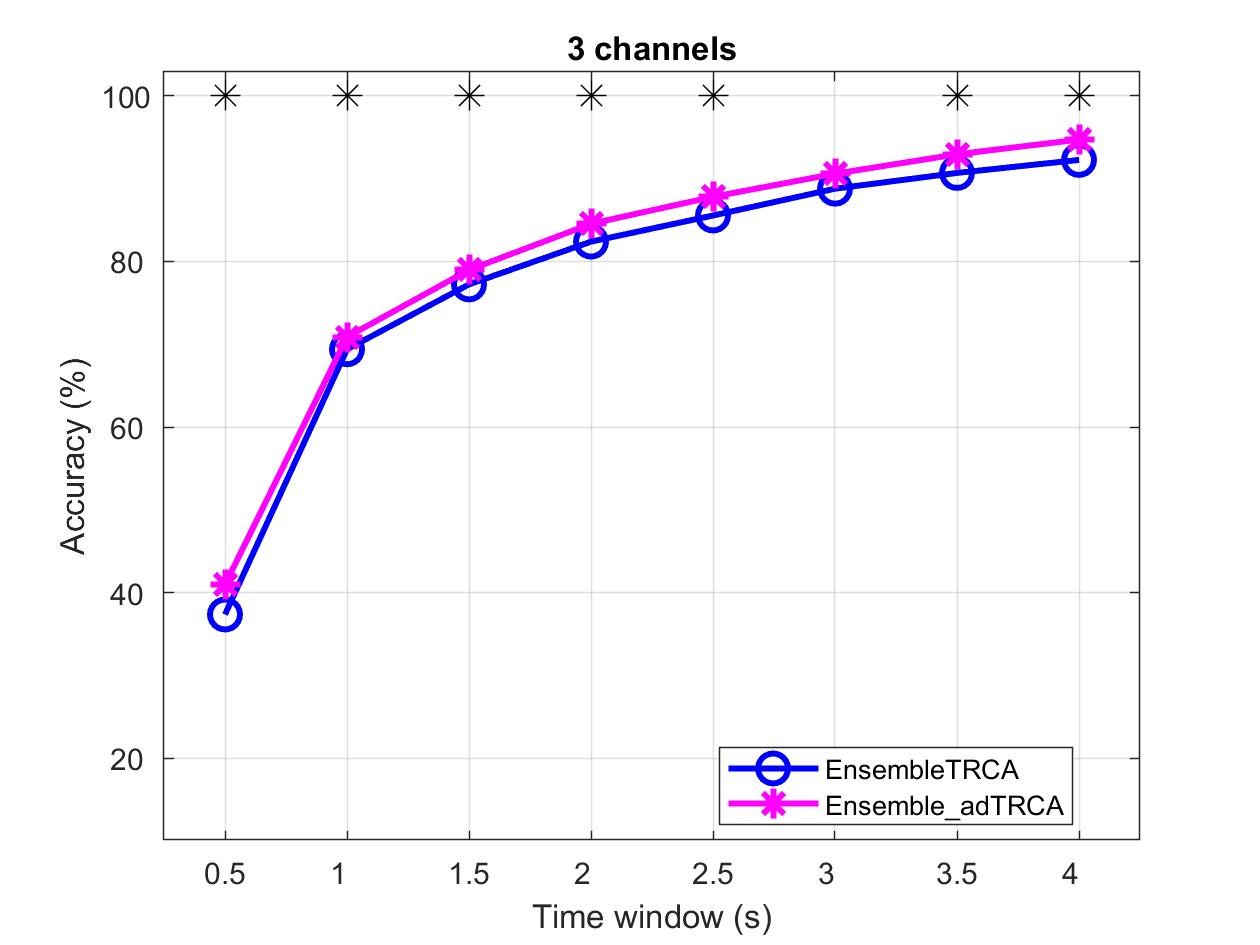} \\
			(a) & (b) \\
			\includegraphics[width=8cm]{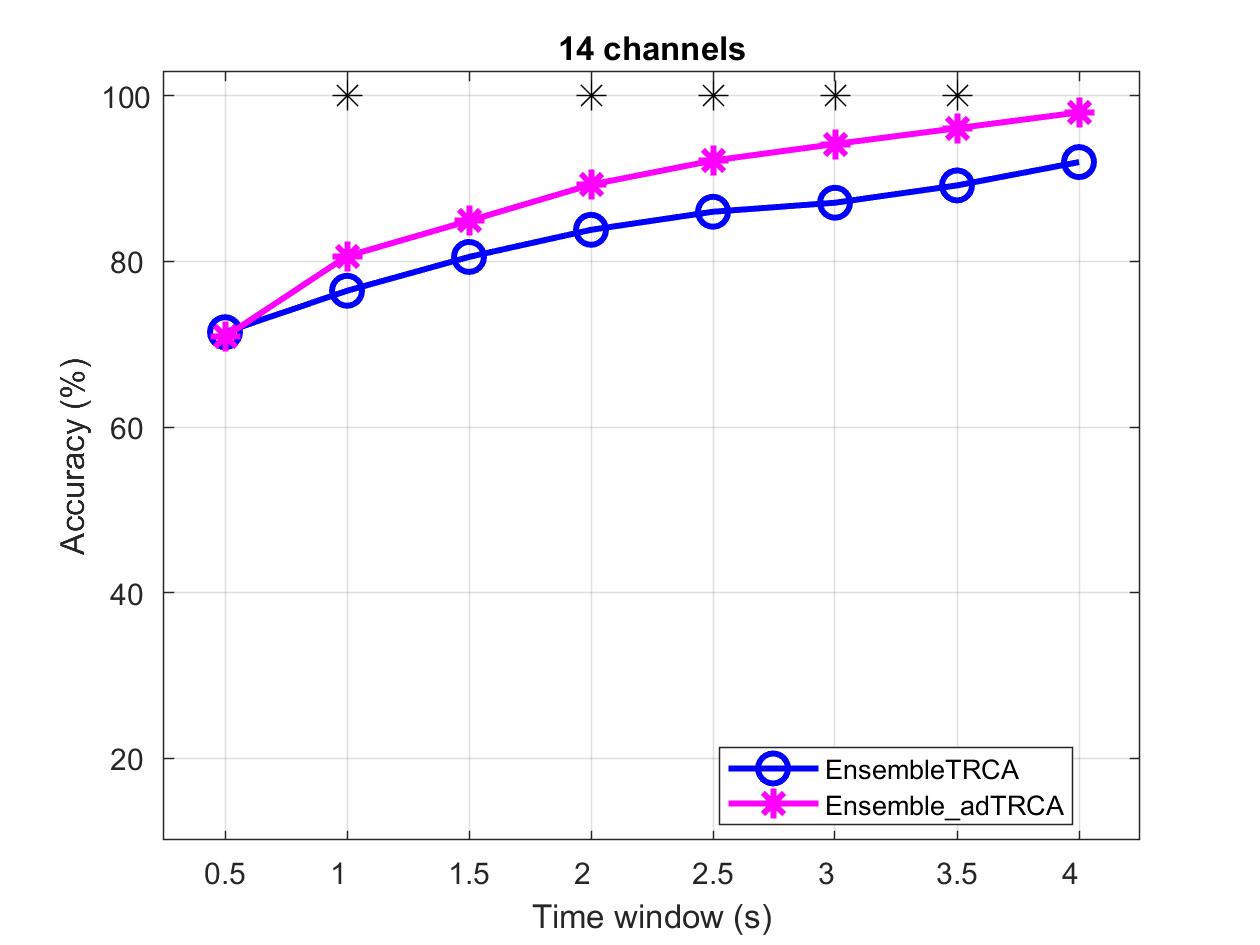} & \includegraphics[width=8cm]{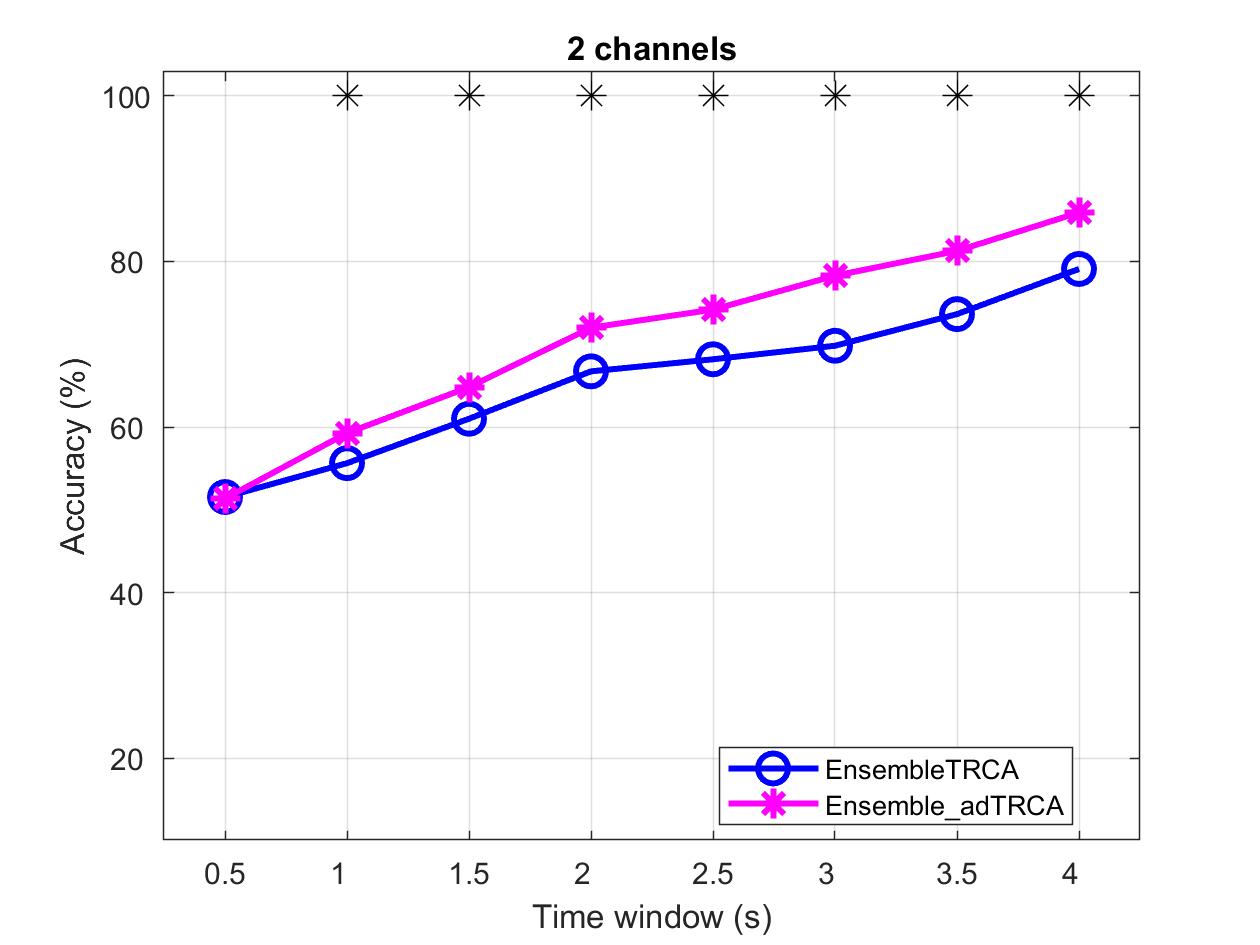}\\  
			(c) & (d) \\
		\end{tabular}
    \endgroup
	\end{center}
	\caption{Average Classification over all subjects by using for the Speller dataset with (a) 9 channels and (b) 3 channels and for the EPOC dataset with (c) 14 channels and (d) 2 channels, respectively.  In both cases, the time window ranges from 0.5s to 4s (0.5s interval). *  indicates statistically significant difference between the two methods using paired sample t-test for Speller dataset and Wilcoxon signed rank test for the EPOC dataset ($p < 0.05$).}
	\label{fig:resEnsemble}
\end{figure*}

In the previous subsection we have presented experiments and we 
performed comparisons using the basic version of our method.
In the current subsection, we provide the obtained results 
of the ensemble version of our method (see \ref{sec:ensemble}),
and also, we perform a comparison with ensemble TRCA
\citep{Nakanishi:2018}. More specifically, in Fig. 	\ref{fig:resEnsemble}
we provide the obtained results for the ensemble TRCA (EnsembleTRCA) and 
ensemble Adaptive TRCA (Ensemble\_adTRCA) methods. The comparison between the two 
methods is performed with respect to the number of channels and the datasets.
For the \textit{Speller} dataset we can observe that in the case of 9 channels
the two methods present similar performance. However, in the case of 3 channels
the Ensemble\_adTRCA method provides significantly better performance than the 
EnsembleTRCA method. Additionally, when we use the \textit{EPOC} dataset,
the Ensemble\_adTRCA method provides significantly better performance, 
either using all 14 channels of the EPOC device or the two channels covering
the occipital lobe. 

\section{Discussion and Conclusions}\label{sec:Conc}
Enhancing the performance of SSVEP recognition is a significant issue for
BCI applications. In this study, we develop a multitask learning scheme 
to strengthen the TRCA method.
The idea behind the proposed learning scheme is to develop a adaptive time - domain
filter which can be used to a more general eigenvalue problem than the corresponding
problem of TRCA method. The proposed method is able to deal with more general 
noises and with reduced number of trials as the experiments have shown. 
However, this increase in performance from our method has an increase also in 
the computation time of the overall procedure since an iterative method is used
to find the adaptive time domain filters. 

A significant part of our study is the use of two SSVEP datasets to evaluate our method.
Most of SSVEP studies use the Speller dataset to evaluate the proposed methods.
However, this dataset was created into a controlled environment with 
high cost EEG equipment which make difficult to replicate the study 
for real BCI applications with low cost equipment and in very noisy environment. 
Hence the aforementioned methods that are evaluated on the Speller dataset tend
to underestimate the noise part of SSVEP EEG trials. An effect which can be observe
by comparing the performance of adTRCA and TRCA on both datasets. We can see
that adTRCA method provides much better performance than the TRCA in the case of EPOC
dataset. In the Speller dataset the adTRCA method has around 1\% better accuracy than
the TRCA, while, in the EPOC dataset this difference is increase to 5\%. 
Additionally, the ensemble version of our method presents better performance than
the ensemble version of TRCA in both datasets, especially when we have 
a limited number of channels. 

The spatial filters and the SSVEP templates play important roles in the target 
recognition methods. When the spatial filters and the SSVEP templates can not 
be accurately computed, e.g. in the case of small calibration data or 
noisy EEG recordings,  the resulting recognition performance will be 
dramatically decreased. 
Hence, to this challenge the key is how to estimate reliable spatial 
filters. 
In this study, we present a novel spatial filtering approach to recognize SSVEP signals.
Our method use the multi-task idea to construct adaptive time - domain filters 
resulting into a generalized eigenvalue problem from where the final spatial 
filters are obtained. 
Extensive experiments, using two SSVEP datasets, have been shown
the usefulness of our method. The proposed method significantly outperformed the TRCA,
the CCA, the MLR and the SRC methods in terms of classification accuracy and ITR.
Finally, future extensions of our approach could include transfer learning approaches utilizing
the data from all subjects to construct the recognition model. 





\section*{Acknowledgments}
This work was part of project NeuroMkt that had been co-financed by the European Regional Development Fund of the European Union and Greek National Funds through the Operational Program Competitiveness, Entrepreneurship and Innovation, under the call RESEARCH CREATE INNOVATE (Project code T2EDK-03661)


\section*{Data Availability Statement}
The datasets that have been used in this study are available in the Internet.
The \textit{Speller} dataset can be found in http://bci.med.tsinghua.edu.cn/download.html. 
The \textit{EPOC} dataset can be found in https://physionet.org/content/mssvepdb/1.0.0/.

 \bibliographystyle{elsarticle-num} 
 \bibliography{adTRCA}





\end{document}